\documentclass[10pt,twocolumn,letterpaper]{article}

\usepackage{iccv}              
\usepackage{times}

\usepackage[accsupp]{axessibility}
\usepackage{pifont}
\usepackage{colortbl}
\usepackage{xcolor}
\usepackage{multirow}
\usepackage{arydshln}
\usepackage{marvosym}
\definecolor{iccvblue}{rgb}{0.21,0.49,0.74}
\usepackage[pagebackref,breaklinks,colorlinks,allcolors=iccvblue]{hyperref}
\def\paperID{8628} 
\def\confName{ICCV}
\def\confYear{2025}

\title{3D Test-time Adaptation via Graph Spectral Driven Point Shift}

\author{Xin Wei\quad Qin Yang\quad Yijie Fang\quad Mingrui Zhu\quad Nannan Wang$\textsuperscript{\Letter}$ \\ Xidian University \\ {\tt\small \{weixin, mrzhu, nnwang\}@xidian.edu.cn} {\tt\small\{qinyang, fangyijie\}@stu.xidian.edu.cn}
}

\begin{document}
\maketitle
\begin{abstract}
While test-time adaptation (TTA) methods effectively address domain shifts by dynamically adapting pre-trained models to target domain data during online inference, their application to 3D point clouds is hindered by their irregular and unordered structure. Current 3D TTA methods often rely on computationally expensive spatial-domain optimizations and may require additional training data. In contrast, we propose Graph Spectral Domain Test-Time Adaptation (GSDTTA), a novel approach for 3D point cloud classification that shifts adaptation to the graph spectral domain, enabling more efficient adaptation by capturing global structural properties with fewer parameters. Point clouds in target domain are represented as outlier-aware graphs and transformed into graph spectral domain by Graph Fourier Transform (GFT). For efficiency, adaptation is performed by optimizing only the lowest 10\% of frequency components, which capture the majority of the point cloud’s energy. An inverse GFT (IGFT) is then applied to reconstruct the adapted point cloud with the graph spectral-driven point shift. This process is enhanced by an eigenmap-guided self-training strategy that iteratively refines both the spectral adjustments and the model parameters. Experimental results and ablation studies on benchmark datasets demonstrate the effectiveness of GSDTTA, outperforming existing TTA methods for 3D point cloud classification.

\end{abstract}    
\section{Introduction}
\label{sec:introduction}

\begin{figure}[t] 
  \centering
  \includegraphics[width=0.47\textwidth]{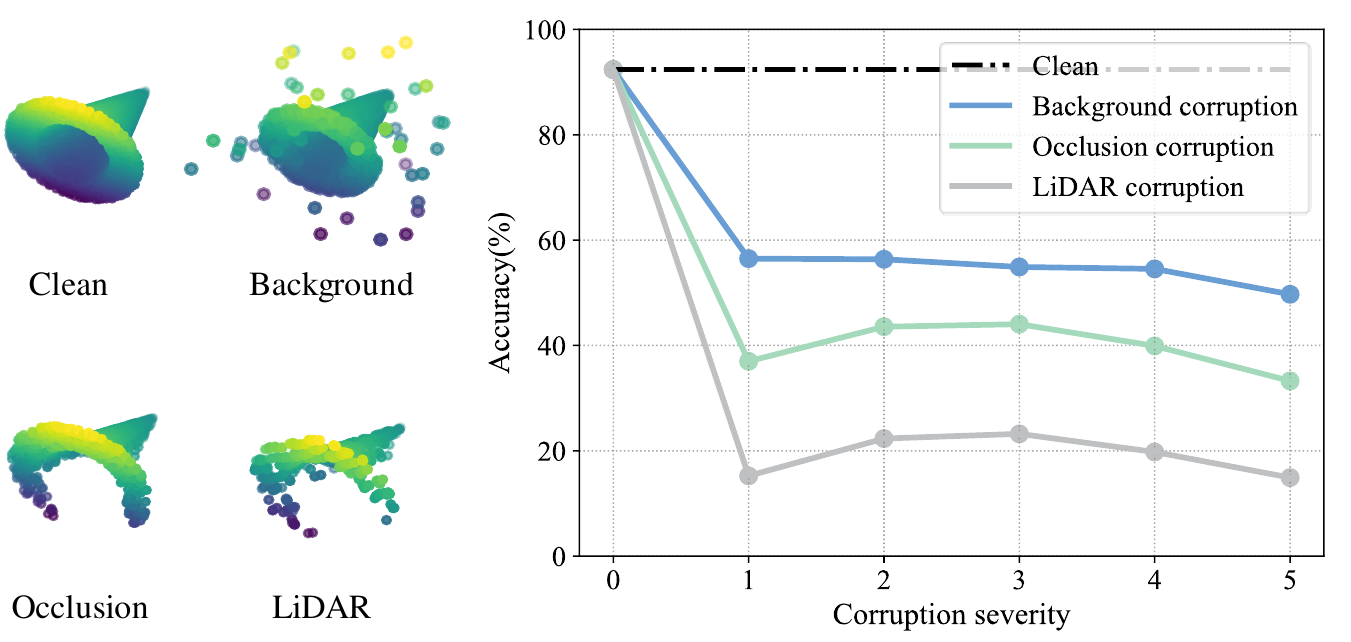} 
  \caption{A point cloud classification model (DGCNN~\cite{dgcnn}), when trained on a clean dataset, suffers a significant performance drop when tested on point clouds with domain shifts.}
  \label{fig:1}
  \vspace{-0.3cm}
\end{figure}

Point cloud classification is a fundamental area within computer vision, with a wide range of applications such as autonomous driving, virtual and augmented reality, and archaeology. Numerous deep models~\cite{pointnet, dgcnn, curvenet, pointnet2, rscnn, pct, simpleview, rpnet, pointcnn, pointmlp, spidercnn, kpconv, pointnext} have recently been developed for point cloud classification, demonstrating impressive performance. However, their success heavily relies on the oversimplified i.i.d. assumption for training and test data, overlooking the challenges of out-of-distribution scenarios that are frequently encountered in real-world applications. As illustrated in Fig.~\ref{fig:1}, a powerful point cloud classification deep model, DGCNN~\cite{dgcnn}, trained on a clean dataset (ModelNet40~\cite{modelnet40}), suffers a significant performance drop (over $35\%$) when tested on point clouds with real-world noises (\textit{e.g.}, Background, Occlusion, and LiDAR corruptions). These corruptions are inevitable, arising from factors such as scene complexity, sensor inaccuracies, and processing errors, which hinders the practical deployment of these models.

Test-time adaptation (TTA) is a technique that enables models trained on the source domain to dynamically adapt to the target domain during online inference, showing significant promise in addressing domain shifts for 2D vision tasks~\cite{bn,tent,t3a,tea,shot,memo,IST,dua,dda,FOA,gda,lame,tda,ttt,ttt++}. At test time, TTA methods typically adapt either model parameters~\cite{bn,tent,t3a,tea,shot,memo,IST,dua,FOA,lame,tda,ttt,ttt++} or test data representations~\cite{dda,gda} to reduce the gap between the training and test data distributions, improving performance on specific test samples. However, the irregular and unordered nature of 3D point clouds limits the direct application of 2D TTA methods to the 3D domain. 

While recent studies have begun exploring TTA for 3D point clouds, the field remains in its early stages. MATE~\cite{mate} introduces an auxiliary self-supervised reconstruction task during source domain training to improve robustness. Alternatively, BFTT3D~\cite{bftt3d} reduces error accumulation during adaptation by combining source-specific features with domain-independent ones from a non-parametric network, but requires extracting feature prototypes from source domain data first. Both CloudFixer~\cite{cloudfixer} and 3DD-TTA~\cite{3ddtta} leverage a diffusion model~\cite{diffusion1,diffusion2,diffusion3}, pre-trained on source data, to repair corrupted point clouds by aligning them with the source distribution. However, both of these works do not strictly adhere to the test-time adaptation setting since they require access to source training data, which is often inaccessible in real-world applications. Moreover, MATE~\cite{mate}, CloudFixer~\cite{cloudfixer} and 3DD-TTA~\cite{3ddtta} rely on the challenging optimization tasks of masked patch reconstruction or per-point transformation learning ($\Delta P\in R^{N\times 3}$ of a point cloud, where $N$ often exceeds $1024$). These high-dimensional optimization problems become particularly challenging when working with limited or streaming test data.

Unlike previous 3DTTA methods that adapt point clouds in the spatial domain, this work shifts the focus to adapting in the graph spectral domain at test time based on two key observations. First, spectral-based point cloud descriptors~\cite{spectral1,spectral2,spectral3,GPS,spectral5,spectral6,spectral7} leverage spectral analysis techniques from graph theory to capture the underlying structure and intrinsic geometric properties of point clouds. These spectral characteristics provide higher-level, global information, which encodes abstract, essential contexts crucial for point cloud recognition. Adjusting low-frequency components in the graph spectral domain requires approximately 90\% fewer parameters than in the spatial domain to control a point cloud's global information, thus reducing optimization complexity with limited test data. Second, graph Laplacian eigenmaps serve as domain-independent descriptors, enabling robust adaptation. These eigenmaps complement the source-specific features extracted from the pre-trained model, which is especially valuable during the early stages of test-time adaptation before the model has fully adjusted to the target domain.

Along this idea, we propose a novel Graph Spectral Domain Test-Time Adaptation (GSDTTA) model for 3D point cloud classification. Given a point cloud classification model pre-trained on source domain data and a batch of test point clouds, GSDTTA adapts the input by operating within the graph spectral domain. Point clouds are represented as outlier-aware graphs and transformed into the spectral domain via the Graph Fourier Transform (GFT). A learnable spectral adjustment is then applied to the low-frequency components of each point cloud. The adjusted GFT coefficients are transformed back to the spatial domain using the inverse GFT (IGFT), resulting in a graph spectral-driven point shift. To optimize this process, we introduce an eigenmap-guided self-training strategy to generate pseudo-labels. This strategy guides the iterative optimization of both the spectral adjustment and the model parameters, progressively refining the adaptation. Extensive experiments on the ModelNet40-C~\cite{modelnetc} and ScanObjectNN-C~\cite{mate} benchmarks confirm the effectiveness of our approach, with GSDTTA achieving significant performance gains over comparable methods.

Our contributions can be summarized as follows. First, we empirically demonstrate that the graph spectral domain of point clouds can capture global structural properties with fewer parameters and provide domain-independent features that facilitate robust cross-domain adaptation at test time. Second, we propose a novel graph spectral domain test-time adaptation model for 3D point cloud classification, featuring an eigenmap-guided self-training strategy for guiding the iterative optimization of spectral adjustments and model parameters. Third, our method achieves state-of-the-art 3D test-time adaptation performance on various benchmarks.

\section{Preliminary and Motivation}

\subsection{Problem Definition}
In the context of 3D test-time adaptation, we consider a point cloud classification model $f_{\theta}(\cdot)$ trained on a source dataset $\mathcal{D}_{S}=\{\mathcal{X},\mathcal{Y}\}$ inaccessible at test time. Each point cloud $X\in \mathcal{X}$ is represented as a set of three-dimensional vectors $X=\{x_i\}_{i=1}^N$, following the distribution ${p(\mathcal{X})}$, and $f_g$ denotes the global deep descriptor of the point cloud extracted by model $f_\theta$. Given an unlabeled target dataset $\mathcal{D}_{T}=\{{\tilde{\mathcal{X}}}\}$, where each point cloud $\tilde{X}\in {\tilde{\mathcal{X}}}$ is drawn from a different distribution $q(\mathcal{X})\neq p(\mathcal{X})$, the objective of test-time adaptation is to enable accurate predictions despite these distribution shifts. Test-time adaptation achieves this by adapting model parameters $\theta$~\cite{shot,mate,IST,3dpl,bftt3d}, the target data $\tilde{X}$~\cite{gda,dda,cloudfixer}, or prompts in transformer-based models~\cite{adapt_p1,adapt_p2,FOA}. Current approaches typically adapt one or a combination of these components in an online or batch-wise manner during inference, without requiring extensive access to target data at each test step.

\begin{figure}[t]
\centering
\includegraphics[width=0.45\textwidth]{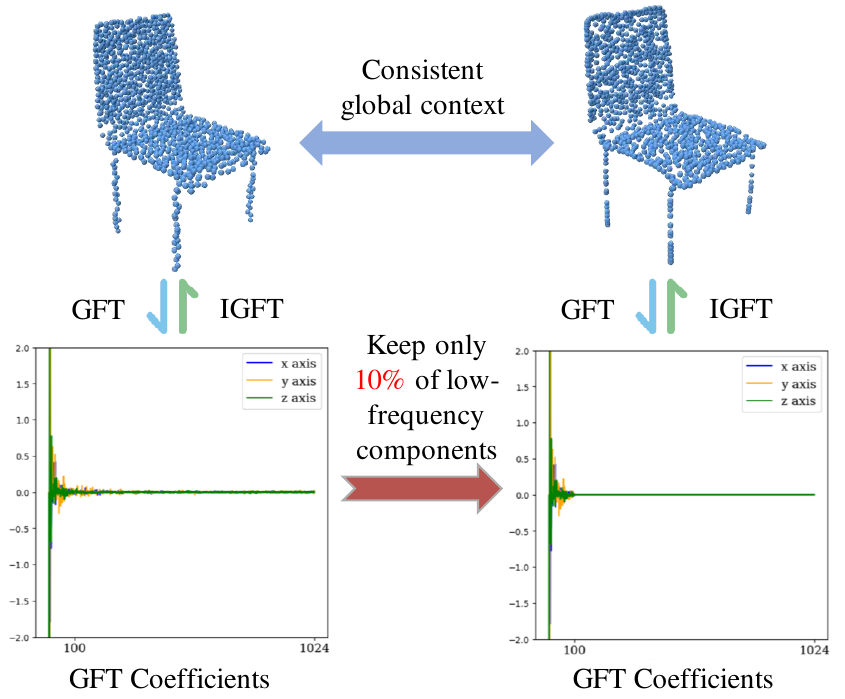}
\caption{Analysis of a chair point cloud's graph spectrum shows that 95\% of the spectral energy is concentrated in the low-frequency components, and that the lowest 10\% of these components are sufficient to reconstruct its global shape.}
\label{fig:2}
\vspace{-0.3cm}
\end{figure}

\subsection{Graph Spectral Analysis for Point Clouds}
Given a point cloud $ {X}\in\mathbb{R}^{N\times 3}$ with ${N}$ points, an undirected graph $G = ( {V}, {A})$ is built with $i$-th node $v_i$ as the $i$-th point $x_i$ in point cloud $X$. The element $A_{ij}$ of adjacency matrix $A\in R^{N\times N}$ is defined as:
\begin{equation}
    {A}_{ij}=\mathbb{I}({x}_{j}\in \mathcal{N}(x_i)),
\end{equation}
where $\mathbb{I}(\cdot)$ is a binary function indicating whether $x_j$ is within the kNN of $x_i$ in spatial domain. The combinatorial graph Laplacian matrix of $G$ is then computed by:
\begin{equation}
     {L}= {D}- {A},
\end{equation}
where $D$ is the diagonal degree matrix with ${D}_{i,i}=\sum_{j=1}^{N}{A}_{ij}$. Since $L$ is a real, symmetric and positive semi-positive matrix, the Laplacian eigenvector matrix $ {U}=[u_1,u_2,...,u_N]$ and the eigenvalue matrix $ {\Lambda}=diag([\lambda_{1},...,\lambda_{N}])$ are computed by eigen decomposition:
\begin{equation}
     {L}= {U} {\Lambda} {U^{T}}.
\end{equation}
In this decomposition, each eigenvector ${u}_{i}$ in ${U}$ is orthogonal to the others, and the eigenvalues $\lambda_{i}$ in $ {\Lambda}$ satisfy the ordering condition $\{\lambda_{1} = 0 \leq ... \leq \lambda_{i} \leq \lambda_{i+1} \leq ... \leq \lambda_{N}\}$. The eigenvalues of a graph are referred to as the graph frequency or spectrum of a point cloud, with larger eigenvalues corresponding to higher graph frequencies. The eigenmaps are subspaces of eigenvectors, constructed by excluding the eigenvector associated with the eigenvalue of $0$, and using the remaining $m$ eigenvectors to embed the graph nodes into an $m$-dimensionals space $E: v_i\rightarrow[u_1(v_i),...,u_m(v_i)]$. We can derive a global spectral descriptor $f_s$ for the point cloud by applying element-wise max-pooling to the embedded features of the graph nodes, which is a simplified variant of the well-known Global Point Signature~\cite{GPS}:
\begin{equation}
    f_s = \operatorname{maxpooling} (E(v_1),...,E(v_m)).
\end{equation}
The spectral coefficients of any vertex $ {v}_{i}$ of $G$ is derived by:
\begin{equation}
     {\hat{v}}_i = \phi _{\operatorname{GFT}}({v}_{i})= {U}^{T} {v}_{i}.
\end{equation}
The inverse Graph Fourier Transform (IGFT) transforms the spectral coefficients to spatial domain:
\begin{equation}
     {v}_i = \phi _{\operatorname{IGFT}}( {\hat{v}}_{i})= {U} {\hat{v}}_{i}.
\end{equation}



\subsection{Motivation on 3D Test-time Adaptation via Graph Spectral Driven Point Shift} \label{subsec:motivation}

As introduced in Sect.~\ref{sec:introduction}, our method adapts point clouds in the graph spectral domain. We motivate this choice by two key properties of this domain, which we then experimentally validate: it efficiently captures global structure with few parameters, and its features are domain-independent, providing a robust complement to potentially source-biased deep features.

The graph spectral domain exhibits remarkable efficiency and invariance. First, as illustrated in Fig.~\ref{fig:2}, it demonstrates strong energy compaction, with about 95\% of a chair point cloud's spectral energy concentrated in its low-frequency components (typically 100 coefficients). This allows us to reconstruct the global context using only the lowest 10\% of coefficients, significantly simplifying the optimization process compared to adapting features in the spatial domain. This is especially beneficial for online or data-limited TTA. Second, the low-frequency eigenmap provides an isometrically invariant shape descriptor that is inherently domain-agnostic. This contrasts sharply with deep features, which often retain source-domain bias, especially early in adaptation. Our ablation studies confirm that augmenting deep features with these stable spectral descriptors enhances adaptation performance.

Capitalizing on these properties, we introduce a graph spectral driven point shift for adaptation. Our method applies a learnable spectral adjustment directly to the low-frequency components of each test point cloud. To optimize this adjustment and the model parameters $\theta$, we employ an eigenmap-guided self-training strategy. This strategy generates high-quality pseudo-labels by forming a convex combination of logits derived from two complementary sources: the global deep descriptors and the robust, domain-independent global spectral descriptors.

\begin{figure*}
  \centering
    \includegraphics[width=0.8\linewidth]{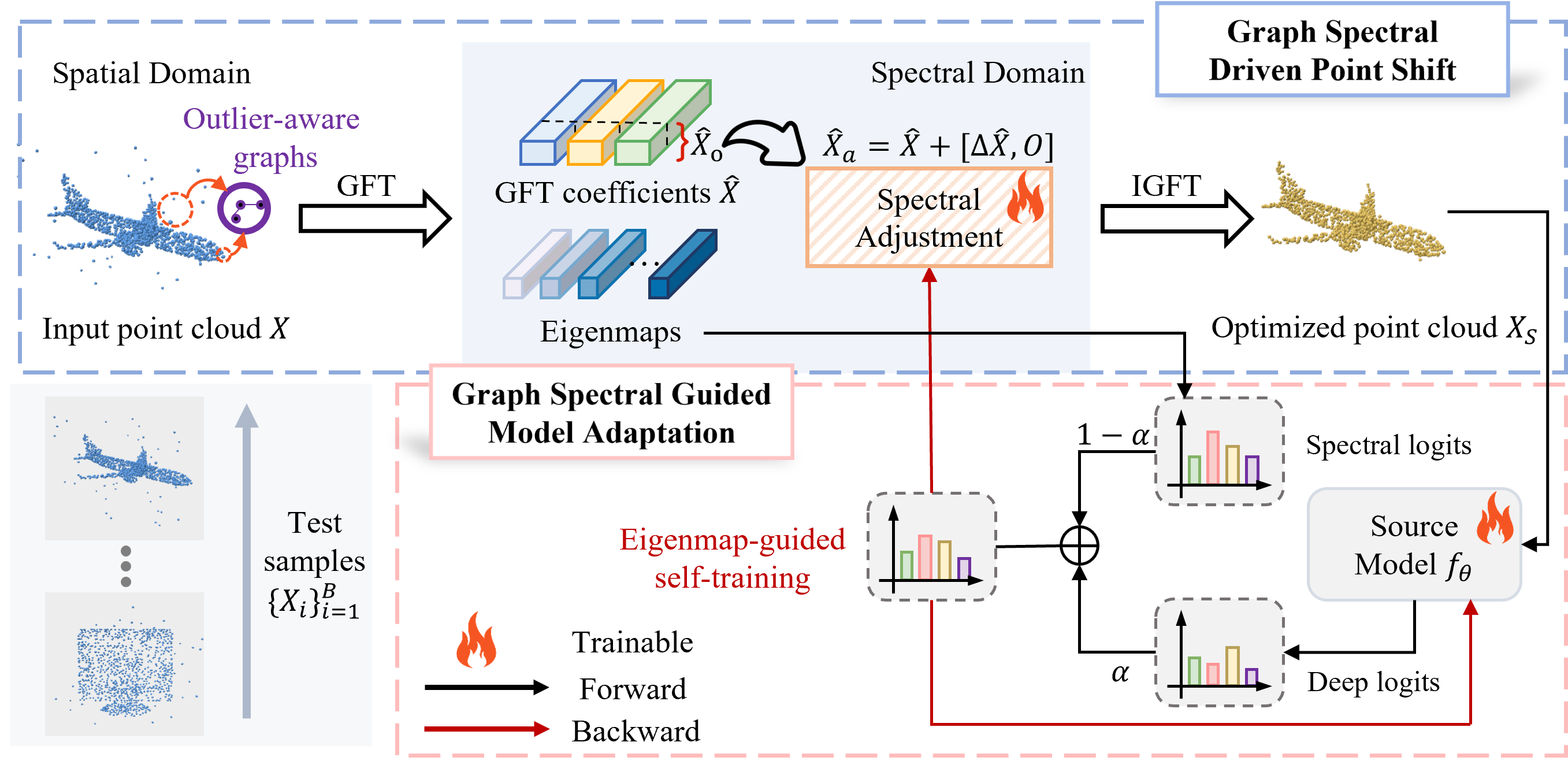}
    \caption{The pipeline of the proposed GSDTTA. Given a batch of test samples $\{X_i\}_{i=1}^B$ and point classification model $f_\theta$ pre-trained on source domain, Graph Spectral Driven Point Shift (GSDPS) and Graph Spectral Guided Model Adaptation (GSGMA) modules iteratively adapt both the point cloud and model in the graph spectral space. This adaptation is achieved by optimizing the spectral adjustment $\Delta \hat{X}$ and model parameters $\theta$ through an eigenmap-guided self-training strategy.}
    \label{fig: framework}
\vspace{-0.3cm}
\end{figure*}

\section{Method}
\label{sec:medthod}
In this section, we introduce the framework for the Graph Spectral Domain Test-Time Adaptation (GSDTTA) model. As shown in Fig.~\ref{fig: framework}, GSDTTA comprises two main components: Graph Spectral Driven Point Shift (GSDPS) and Graph Spectral Guided Model Adaptation (GSGMA). The input point cloud $X$ and the point cloud classification model $f_\theta$ are iteratively adapted by GSDPS and GSGMA, progressively refining the adaptation process at test time.

\subsection{Graph Spectral Driven Point Shift}\label{subsec:method_input}
In the Graph Spectral Driven Point Shift (GSDPS) model, each input point cloud is adapted through a point shift derived from graph spectral adjustment. A point cloud is initially constructed as an outlier-aware graph, transformed into the graph spectral domain, and adjusted in its low-frequency components via a spectral adjustment. The adjusted spectral representation is then converted back to the spatial domain, resulting in a point cloud with a graph spectral driven point shift. The spectral adjustment is optimized through an eigenmap-guided self-training strategy.

\vspace{-0.3cm}

\paragraph{From point clouds to outlier-aware graphs.} Given a point cloud ${X}\in\mathbb{R}^{N\times 3}$, we first construct a graph $G_{o}=\{V,A_{o}\}$ upon $X$. We use Radial Basis Function (RBF) as the weight function for the edges between points $x_i$ and $x_j$:
\begin{equation}
    w_{ij} = exp(-\frac{d^2(x_i,x_j)}{2\delta^2}),
\end{equation}
where $d(\cdot,\cdot)$ denotes the squared Euclidean distance between two vertices and $\delta$ is a hyperparameter. The element of the adjacency matrix $A$ is then given by:
\begin{equation}
    {A}_{ij}=w_{ij}\cdot \mathbb{I}({x}_{j}\in \mathcal{N}(x_i)),
\end{equation}
where $\mathbb{I}(\cdot)$ is an indicator function where edges are only kept if $x_j$ is within the $k$-NN neighborhood $\mathcal{N}(x_i)$.

Since spectral analysis of point cloud graphs can be sensitive to outliers, we leverage the fact that outliers are often far from inlier points. As a result, the degree of outlier vertices—defined as the sum of weights on all adjacent edges—tends to be significantly lower than that of inlier points. Erroneous points can therefore be removed by eliminating vertices with degrees below a threshold $\tau$. The element of final adjacency matrix $A_{o}$ is defined as:
\begin{equation}
    {A}_{ij}^{o}=w_{ij} \cdot \mathbb{I}({x}_{j}\in \mathcal{N}(x_i))\cdot \mathbb{I}(\sum_{j=1}^N A_{ij} >\tau),
\end{equation}

The threshold $\tau$ is calculated as $\gamma$ times the average ${k}$-nearest neighbor distance across the entire point cloud, providing a global measure of point dispersion:
\begin{equation}
    \tau=\frac{\gamma}{Nk}\sum_{i=1}^N\sum_{j=1}^N A_{ij}.
    \label{eq:gamma}
\end{equation}

\vspace{-0.3cm}

\paragraph{Spectral adjustment driven point shift.} The Laplacian matrix ${L_o}$ of the outlier-aware graph $G_o$ is computed by $L_o=D_o-A_o$ where $D_o$ is the degree matrix, and then decomposed to obtain the eigenvector matrix ${U_o}$ by solving $L_o=U_o\Lambda_o U_o^T$. These eigenvectors are used to compute the GFT coefficients as follows:
\begin{equation}
    {\hat{X}} = {U}_o^{T}{X},
\end{equation}
where ${\hat{X}} \in\mathbb{R}^{N\times 3}$ represents the transformed coefficients of signal in the three axes. Then a learnable spectral adjustment $\Delta \hat{X}\in R^{M\times3}, M<<N$ is deployed to adjust the coefficients ${\hat{X}}$ of low frequency:
\begin{equation}
    {\hat{X}}_a = {\hat{X}} + \Delta \hat{X}^{'},
    \label{eq:adjust}
\end{equation}
where $\Delta \hat{X}^{'} = [\Delta \hat{X}, O]\in R^{N\times3}$ is defined as the concatenation of $\Delta \hat{X}\in R^{M\times3}$ and a zero matrix $O\in R^{(N-M)\times3}$. Finally, the adjusted spectral representation ${\hat{X}}_a$ is converted back to the spatial domain to obtain the point cloud with a graph spectral driven point shift:
\begin{equation}
    {X}_s = {U_o}{\hat{X}}_a.
\end{equation}
The spectral adjustment $\Delta \hat{X}$ will be optimized automatically according to the objective function that will introduced in following section.

\vspace{-0.3cm}

\paragraph{Optimizing spectral adjustment by Eigenmap-guided self-training.} To optimize the spectral adjustment $\Delta \hat{X}$, as discussed in Sect.~\ref{subsec:motivation}, we propose an eigenmap-guided self-training strategy to generate pseudo-labels for self-supervised training. Given a batch of point clouds $\{X_i\}_{i=1}^B$ with global deep descriptors $\{f_d^i\}_{i=1}^B$ and global spectral descriptors $\{f_s^i\}_{i=1}^B$, the centroids $q_d^c$ for $c$-th class in the global deep descriptor space are defined as:
\begin{equation}
q_d^c = \frac{\sum_{i=1}^B (f_\theta(X_i))_c f_d^i}{\sum_{i=1}^B (f_\theta(X_i))_c},
\end{equation}
where $(f_\theta(X_i))_c\in R$ is the class probability for the $c$-th class of the target sample $X_i$. Similarly, the centroids $q_s^c$ in the global spectral descriptor space are defined as:
\begin{equation}
q_s^c = \frac{\sum_{i=1}^B (f_\theta(X_i))_c f_s^i}{\sum_{i=1}^B (f_\theta(X_i))_c}.
\end{equation}
The centroids $q_d^c$ and $q_s^c$ serve as soft cluster assignments for class $c$, providing robust representations to guide adaptation. The final pseudo-label $\hat{y}_i$ for test sample $X_i$ is generated as a convex combination of the class probabilities of $f_d^i$ and $f_s^i$:
\begin{equation}
    \hat{y}_i = \arg \min_c (\alpha\frac{(f_d^i)^Tq_d^c}{\Vert f_d^i\Vert \Vert q_d^c\Vert}+(1-\alpha)\frac{(f_s^i)^T q_s^c}{\Vert f_s^i\Vert \Vert q_s^c\Vert}),
    \label{eq:combine}
\end{equation}
where $\alpha$ is a weight factor to balance the two terms. The overall input adaptation objective is:
\begin{equation}
    \arg \min_{\Delta \hat{X}} \mathcal{L}_{IA} = \arg \min_{\Delta \hat{X}} ( \mathcal{L}_{pl} + \beta_1({\mathcal{L}_{ent}} + {\mathcal{L}_{div}}) + {\beta_2 \mathcal{L}_{cd}}),
    \label{eq:IA}
\end{equation}
where $\mathcal{L}_{pl}=CE(f_\theta(X_s),\hat{y})$ is the cross entropy loss. $\mathcal{L}_{ent}$ denotes the entropy loss as $-\sum_{c=1}^C f_\theta(X_s) \log f_\theta(X_s)$, which encourages the model to make more confident predictions on the optimized point cloud. Divergency loss $\mathcal{L}_{div}=\sum_{c=1}^C g_c\log(g_c)$, where $g_c=\frac{1}{B} \sum_{i=1}^B (f_\theta(X_i))_c$, promotes diversity in the outputs while ensuring individual certainty. Together, $\mathcal{L}_{ent}$ and $\mathcal{L}_{div}$ form an information maximization loss~\cite{im1,im2}. $\mathcal{L}_{cd}$ is the single direction Chamfer Distance from input point cloud $X$ to adapted point cloud $X_s$, encouraging $X$ to be a part of $X_s$. $\beta_1$ and $\beta_2$ are weight factors controlling the relative contributions of different losses.

\subsection{Graph Spectral Guided Model adaptation}\label{subsec:method_model}
To optimize the model's adaptation to the target domain, we apply graph spectral-guided model adaptation to adjust the parameters 
$\theta$ of the point cloud classification model $f_\theta$. The objective function of model adaptation is:
\begin{equation}
    \arg \min_{\theta} \mathcal{L}_{MA} = \arg \min_{\theta} (\mathcal{L}_{pl} + \beta_3({\mathcal{L}_{ent}} + {\mathcal{L}_ {div}))},
    \label{eq:MA}
\end{equation}
where $\mathcal{L}_{pl}$, $\mathcal{L}_{ent}$, and $\mathcal{L}_{div}$ are the same losses defined in the input adaptation step. $\beta_3$ is a weight factor.
\begin{table*}[!t]
    \belowrulesep=0pt
    \aboverulesep=-2pt
    \centering
    \renewcommand{\arraystretch}{1.05}
    \resizebox{\linewidth}{!}{ 
    \begin{tabular}{c|c|cccccccccccccccccc|c}
        \toprule
        \textbf{Backbone} & \textbf{Method} & \textbf{uniform} & \textbf{gaussian} & \textbf{background} & \textbf{impulse} & \textbf{upsampling} & \textbf{rbf} & \textbf{rbf-inv} & \textbf{den-dec} & \textbf{dens-inc} & \textbf{shear} & \textbf{rot} & \textbf{cut} & \textbf{distort} & \textbf{occlusion} & \textbf{lidar} & \textbf{Mean}
         \\
        \toprule
        \multirow{10}{*}{DGCNN \cite{dgcnn}} & Source-only & 79.57 & 72.16 & 49.71 & 64.70 & 67.99 & 78.03 & 80.47 & 73.05 & 82.73 & 85.08 & 59.52 & 75.81 & 80.71 & 33.26 & 14.91 & 66.51 \\
        & BN \cite{bn} & 84.48 & 82.82 & 48.01 & 81.32 & 79.98 & 81.68 & 83.22 & 78.36 & 86.10 & 85.53 & 72.24 & 81.23 & 82.90 & 41.12 & 31.11 & 73.34 \\
        & PL \cite{3dpl} & 85.29 & 83.67 & 65.64 & 83.46 & 80.51 & 82.77 & 84.35 & 79.29 & 85.89 & 84.88 & 74.39 & 81.96 & 82.86 & 38.69 & 31.68 & 75.02 \\
        & DUA \cite{dua} & 84.48 & 83.10 & 50.97 & 81.80 & 79.94 & 82.00 & 83.26 & 79.29 & 86.22 & 85.94 & 71.96 & 81.56 & 82.49 & 42.09 & 32.09 & 73.81 \\
        & TENT \cite{tent} & 86.02 & 84.88 & 60.65 & 83.54 & 82.73 & \underline{83.18} & 84.76 & \underline{80.83} & \textbf{87.19} & \underline{86.83} & 75.32 & \underline{82.98} & 83.46 & 42.94 & 33.38 & 75.91 \\
        & SHOT \cite{shot} & 85.69 & 83.95 & \underline{81.40} & 84.19 & 82.21 & 82.86 & 83.75 & 79.86 & 85.25 & 84.35 & \underline{77.95} & 82.41 & \underline{83.95} & \textbf{48.46} & \underline{34.11} & \underline{77.36} \\
        & BFTT3D \cite{bftt3d} & 78.47 & 71.45 & 46.85 & 66.75 & 70.87 & 75.69 & 78.43 & 73.12 & 81.90 & 82.35 & 56.45 & 75.49 & 78.43 & 34.80 & 16.75 & 65.86 \\
        & CloudFixer \cite{cloudfixer} & \textbf{89.95} & \textbf{90.15} & 74.55 & \textbf{90.11} & \underline{85.98} & 82.13 & \underline{84.81} & 73.46 & 84.76 & 82.70 & 77.67 & 76.74 & 81.65 & 35.94 & \textbf{37.48} & 76.54 \\
        & 3DD-TTA \cite{3ddtta} & 85.58&	84.00&	62.48&	76.13	&\textbf{88.41}&	78.36&	80.59&	73.30	&84.85&	82.74	&59.93&	72.77&	79.34&	38.41&	28.48&	71.69\\
        \rowcolor{gray!12}& GSDTTA (ours) & \underline{87.88}&	\underline{86.26}&	\textbf{88.57}&	\underline{86.91}&	84.12&	\textbf{85.05}&	\textbf{86.18}&	\textbf{82.46}&	\underline{86.83}&	\textbf{87.76}&	\textbf{78.53}& \textbf{84.20}&\textbf{84.44}&	\underline{45.38}&	31.52	&\textbf{79.07}\\
        \hline
        \multirow{9}{*}{CurveNet \cite{curvenet}} & Source-only & 88.13	&84.76	&15.56&	65.56&	89.10	&85.49&	86.18&	78.81&	87.97&	87.20&	70.83&	78.44&	86.63&	36.06&	29.98&	71.38 \\
        & BN \cite{bn} & 89.38&	87.40	&36.26&	80.35&	89.34	&86.67&	87.76&	84.04	&88.98&	88.01	&78.69&	84.64&	87.12&	47.20&	45.91&	77.45 \\
        & PL \cite{3dpl} & 89.26&	88.45&	36.43	&83.87&	88.86&	\underline{87.68}&	\textbf{88.94}&	85.29	&\underline{89.30}	&\underline{88.74}&	82.70&	\underline{87.03}&	\underline{87.44}&	47.33&	\underline{47.12}&	78.56 \\
        & DUA \cite{dua} & 89.30	&87.40&	36.26&	80.27	&89.38&	86.63&	87.68&	83.91&	89.02	&87.97	&78.48	&84.56&	87.12&	47.12&	45.98&	77.41 \\
        & TENT \cite{tent} & 89.42&	87.56	&36.63	&81.48&	89.47&	87.28&	87.68&	84.48	&88.94	&88.25&	79.42&	85.41&	87.12&	48.14&	46.88&	77.88 \\
        & SHOT \cite{shot} & 87.56&	87.40&	\underline{66.49}&	86.18&	83.83	&87.40&	88.01&	\textbf{85.78}	&87.28	&87.48&	\textbf{83.95}&	86.30&	86.35&	\textbf{58.63}&	\textbf{56.04}&	\underline{81.24 }\\
        & BFTT3D \cite{bftt3d} & 85.63	&81.86&	16.07&	66.64&	89.26&	84.60&	84.93&	79.25&	87.79	&86.36	&68.18	&79.45	&86.15&	37.13&	30.96&	70.95 \\
        & CloudFixer \cite{cloudfixer} &\textbf{90.14}&	\textbf{89.98}&	66.07&	\textbf{90.06}&	\textbf{90.87}&	85.59&	87.30&	76.18&	86.20&	85.06&	81.86&	78.61&	84.86&	37.13&	38.76&	77.91 \\
        \rowcolor{gray!12}& GSDTTA (ours) &  \underline{89.74}	&\underline{89.30}	&\textbf{87.84}&	\underline{87.88}&	\underline{89.87}&	\textbf{88.53}&	\underline{88.61}&	\underline{85.66}	&\textbf{89.55}&	\textbf{89.22}&	\underline{82.90}&	\textbf{87.20}&	\textbf{87.97}&	\underline{50.73}	&44.45 &\textbf{82.63}\\
        \hline
        \multirow{9}{*}{PointNeXt \cite{pointnext}} & Source-only & 69.12	&57.86	&50.81	&70.62	&77.03	&75.04	&77.55	&86.18	&87.84	&79.01	&42.50	&85.82	&76.46	&41.05	&27.96	&66.99\\ 
       &BN \cite{bn}& 86.63&	84.81&	78.69&	87.03	&88.13	&84.16&	85.78&	\underline{89.71}&	\underline{90.92}&	84.68&	70.10&	\underline{89.55}&	83.43&	51.18&	45.54&	80.02\\
       &PL \cite{3dpl}& 87.15&	85.13	&78.89&	87.93&	86.79&	85.01	&\textbf{86.79}	&89.10&	90.03&	\underline{86.06}&	77.76&	88.70&	\underline{84.85}&	51.62&	46.35&	80.81 \\
       &DUA \cite{dua}& 87.32&	85.37&	79.78	&87.88&	88.45&	84.72&	86.18&	\textbf{89.91}&	90.76&	84.72&	72.16&	89.34&	83.58&	51.94&	46.39&	80.57 \\
       &TENT \cite{tent}& \underline{87.80}	&86.43&	80.43&	88.25&	\underline{88.70}&	\underline{85.05}&	86.30&	89.38&	\textbf{91.09}&	85.37&	74.59&	\textbf{89.63}&	84.32&	51.90&	\underline{46.92}&	\underline{81.08} \\
       &SHOT \cite{shot}& 86.39	&85.74&	\underline{81.44}&	85.41&	81.28&	84.20&	85.70&	87.96&	88.65&	83.43&	\underline{79.54}&	88.01&	84.68&	\textbf{55.43}&	\textbf{49.03}&	80.46 \\
       &BFTT3D \cite{bftt3d}& 70.17	&61.24	&54.22	&73.13	&78.25	&75.81	&77.72	&87.18	&88.76	&80.36	&43.71	&86.97	&77.52	&42.45	&28.41	&68.39 \\
       &CloudFixer \cite{cloudfixer}& \textbf{87.91}	&\textbf{88.32}	&79.28&	\underline{88.36}&	\textbf{88.98} &	80.26&	82.32&	80.47&	82.32&	76.69&	65.42&	83.18&	83.06&	38.32	&35.73	&76.04 \\
        \rowcolor{gray!12}&GSDTTA (ours)&  87.48	&\underline{86.71}&	\textbf{91.29} &	\textbf{88.81} &	88.29&	\textbf{85.82}&	\underline{86.75}&	89.06&	90.48&	\textbf{86.06}&	\textbf{80.06}&	89.02&	\textbf{85.98}&	\underline{55.06}	&46.84	&\textbf{82.51}\\
        \bottomrule          
    \end{tabular}}
        \caption{Classification accuracy (\%) is provided for each distribution shift in the ModelNet40-C dataset \cite{modelnetc}. These results reflect the performance of backbone models trained on ModelNet40 \cite{modelnet40} and adapted to the corrupted dataset using a batch size of $32$.  Source-only indicates the accuracy achieved on corrupted test data without applying any adaptation method. The mean accuracy scores are reported, with the highest values highlighted in bold and the second highest underlined.}
        \label{tab:modelnetc}
        \vspace{-0.3cm}
\end{table*}

In GSDTTA, the input and model adaptations are optimized in an iterative manner to improve the point cloud classification model’s performance under domain shifts. This process alternates between two steps: optimizing the spectral adjustment $\Delta\hat{X}$ for input adaptation and updating the model parameters $\theta$ for model adaptation. By iteratively refining both the point clouds and the model parameters, GSDTTA achieves better alignment between the test data and the pre-trained model, leading to enhanced classification accuracy when faced with challenging domain shifts.

\section{Experiment}
In this section, experiments are conducted on ModelNet40-C and ScanObjectNN-C benchmarks to verify the efficacy of the proposed GSDTTA. 

\subsection{Datasets}
\paragraph{ModelNet40-C.} ModelNet40\cite{modelnet40} dataset is a 3D point cloud classification benchmark containing 12,311 shapes across 40 categories (9,843 for training, 2,468 for testing). From this, ModelNet40-C\cite{modelnetc} benchmark was created to evaluate model robustness. It augments the original test set with 15 common, realistic corruptions organized into three categories: transformations, noise, and density variations. These corruptions simulate real-world distributional shifts, providing a rigorous test of model reliability. For further details, please refer to~\cite{modelnetc}. \vspace{-0.3cm}

\paragraph{ScanObjectNN-C.} ScanObjectNN~\cite{scanobjectnn} is a real-world point cloud classification dataset derived from scanned indoor scenes, comprising 15 object categories with 2,309 training samples and 581 testing samples. For consistent robustness evaluation, the ScanObjectNN test set is augmented with the same 15 corruption types applied in ModelNet40-C, forming the ScanObjectNN-C dataset~\cite{mate}.


\subsection{Implementation Details}
\label{sec:implement}
For experiments on the three benchmarks above, we use  DGCNN~\cite{dgcnn}, CurveNet~\cite{curvenet}, and PointNeXt~\cite{pointnext} as the point cloud classification model $f_\theta$ across all comparable methods. For fair comparison, all methods use the same pre-trained weights of backbone networks for each dataset. We report results obtained by running the published code for each method, with detailed implementation information provided in the supplementary material.

As discussed in Sect.~\ref{subsec:method_model}, for GSDTTA, input and model adaptations are optimized iteratively to enhance the robustness and performance of the point cloud classification model under domain shifts. For each batch of test data, GSDTTA first adapts the input point cloud over $4$ steps, followed by $1$ step of model adaptation, repeating this cycle for a total of $10$ steps. The objective function in Eqn.~\ref{eq:IA} and Eqn.~\ref{eq:MA} for both stages is optimized using the AdamW optimizer~\cite{adamw}, with a learning rate of $0.0001$ and batch size of $32$. The parameters $k$ for $k$-NN,$\delta$ and $\gamma$ for constructing the outlier-aware graph in Sect.~\ref{subsec:method_input} are set to $10$, $0.1$ and $0.6$, respectively. In the graph spectral domain, the number of frequency components is set to $M=100$, as defined in Eqn.~\ref{eq:adjust}. The weight factor $\alpha$ in Eqn.~\ref{eq:combine}, $\beta_1$, $\beta_2$ in Eqn.~\ref{eq:IA} for input adaptation, $\beta_3$ for model adaptation in Eqn.~\ref{eq:MA} are set to $0.5$, $0.3$, $1000$, $3$ respectively. All experiments are conducted on a single NVIDIA RTX 3090 GPU.

\begin{table*}[!ht]
    \belowrulesep=0pt
    \aboverulesep=-2pt
    \centering
    \renewcommand{\arraystretch}{1.05}
    \resizebox{\linewidth}{!}{ 
    \begin{tabular}{c|c|cccccccccccccccccc|c}
        \toprule
        \textbf{Backbone} & \textbf{Method} & \textbf{uniform} & \textbf{gaussian} & \textbf{background} & \textbf{impulse} & \textbf{upsampling} & \textbf{rbf} & \textbf{rbf-inv} & \textbf{den-dec} & \textbf{dens-inc} & \textbf{shear} & \textbf{rot} & \textbf{cut} & \textbf{distort} & \textbf{occlusion} & \textbf{lidar} & \textbf{Mean} \\
        \toprule
        \multirow{10}{*}{DGCNN \cite{dgcnn}} & Source-only & 46.99&	44.75&	40.96&	65.75&	56.63&	70.40&	71.94	&67.64	&73.32	&72.63	&61.79	&68.33&	\underline{73.32}&	10.67&	\underline{10.67}	&55.72 \\
        & BN \cite{bn} & 56.28	&52.66	&25.47	&67.81&	62.13	&\textbf{71.42}	&\underline{73.67}&	69.01	&74.35&	73.67&	66.78&	70.56	&\textbf{73.67}	&9.98	&9.81	&57.15 \\
        & PL \cite{3dpl} & 60.75&	55.93&	21.51&	70.39&	\textbf{67.12}&	69.87&	72.11&	69.53&	73.32&	72.28&	\underline{66.95}&	\textbf{71.42}&	72.46&	\textbf{11.18}	&10.32&	57.68 \\
        & DUA \cite{dua} & 57.31&	53.87&	22.37&	68.84&	64.19	&70.91	&72.81&	70.39&	\underline{74.69} & \underline{74.52}&	\textbf{67.12}&	70.74	&\underline{73.32}	&10.67	&10.32	&57.47 \\
        & TENT \cite{tent} & 60.24	&54.73	&19.27&	70.39&	65.57&	70.91&	72.11&	68.15&	74.35&	73.14&	66.26&	\underline{70.91}&	73.14&	\underline{10.84}&	9.63&	57.31 \\
        & SHOT \cite{shot} & 59.89	&\underline{59.04}&	17.21&	70.05&	\underline{68.15}&	69.01&	70.22&	67.98&	70.39&	69.53&	65.40	&67.81&	69.53&	10.67	&9.98&	56.32 \\
        & BFTT3D \cite{bftt3d} & 48.96&	48.96&	41.32&	66.84&	58.68&	71.18&	72.57&	67.88&	72.05&	73.96&	61.98	&68.75	&72.92&	10.24&	\textbf{10.76}&	56.47 \\
        & CloudFixer \cite{cloudfixer} & \textbf{71.70}	&\textbf{68.92}	&46.18	&\textbf{75.00}&	\textbf{72.92}&	70.14&	72.05&	66.32&	73.09	&72.40&	61.46&	69.79&	73.14&	9.55&	8.33&	\underline{60.73} \\
        & 3DD-TTA \cite{3ddtta}& 58.52&	54.04&	\underline{46.64}&	65.75&	62.82&	67.13&	70.91&	\underline{69.71}&	74.01	&71.08	&58.69	&68.67	&70.91	&8.95&	8.43&	57.08 \\
        \rowcolor{gray!12}& GSDTTA (ours) &  \underline{63.17}&	58.52	&\textbf{69.54}&	\underline{73.67}&	66.09	&\underline{71.26}&	\textbf{74.01}&	\textbf{70.74}&	\textbf{75.04}&	\textbf{74.87}&	66.61&	69.02&	\textbf{73.67}&	10.67&	10.51&	\textbf{61.83}\\
        \hline
        \multirow{9}{*}{CurveNet \cite{curvenet}} & Source-only &44.75&	37.35&	24.96&	40.62&	51.29	&71.77&	74.35&	68.68&	76.42&	74.53&	65.92	&70.91	&73.84&	10.33&	10.15&	53.06 \\
        & BN \cite{bn} & 56.28&	50.26&	27.19&	54.04&	62.99&\textbf{72.29}&	75.22&	71.77&	\underline{76.76}	&74.87	&70.57&	71.94&	\textbf{75.22}&	10.67&	9.81&	57.33 \\
        & PL \cite{3dpl} & 62.99	&52.67&	30.12&	58.86&	62.65	&70.22&	74.01&	72.29	&75.04	&75.04&	71.26&	71.08&	72.98&	10.15&	9.12&	57.90 \\
        & DUA \cite{dua} & 60.58	&55.07&	28.05&	57.83&	64.37&	70.57&	\underline{75.73}&	\textbf{73.15}&	76.42&	\underline{75.56}&	71.26&	71.94&	\underline{74.87}&	10.50&	9.29	&58.35 \\
        & TENT \cite{tent} & 62.13	&55.94&	29.43&	58.52&	64.89&	70.91&	\underline{75.73}&	\underline{72.98}	&\underline{76.76}	&\underline{75.56}&	\underline{71.94}&	71.43&	74.35&	\underline{10.84}&	9.29&	58.71 \\
        & SHOT \cite{shot} & \underline{65.75}&	\underline{59.04}&	22.38&	65.57&	61.96&	70.91&	73.49&	70.91	&74.01	&74.01	&70.05&	\underline{72.12}&	72.81&	9.12&	\textbf{10.50}&	58.17 \\
        & BFTT3D \cite{bftt3d} & 50.87&	42.01&	25.35&	43.06&	54.51&	69.62&	73.44&	64.76&	69.62	&71.53&	66.32&	62.85&	73.44&	\textbf{10.94}&	\underline{10.24}&	52.57 \\
        & CloudFixer \cite{cloudfixer} &\textbf{69.79}&	\textbf{68.58}&	\underline{31.77}&	\textbf{75.00}&	\textbf{70.14}&	67.71&	72.74&	63.02&	73.44&	70.14&	64.76&	68.92&	71.88&	10.59&	5.90&	\underline{58.96}\\
        \rowcolor{gray!12}& GSDTTA (ours) &  64.37&	58.00&	\textbf{67.13}&	\underline{69.71}&	\underline{66.78}&	\underline{71.94}&	\textbf{75.90}&	70.05&	\textbf{77.12}&	\textbf{76.08}&	\textbf{72.12}&	\textbf{73.67}&	\textbf{75.22}&	10.67&	10.15 &\textbf{62.59}\\
        \hline
        \multirow{9}{*}{PointNeXt \cite{pointnext}} & Source-only & 32.70&	23.58&	39.41&	46.82&	44.06&	68.67&	69.36&	\underline{73.49}&	74.53&	70.40&	55.25&	73.32&	71.43&	9.12&	7.75&	50.66\\
        &BN \cite{bn}& 46.64&	38.73	&43.55&	59.21&	58.52&	72.46	&\textbf{74.01}	&\underline{73.49}	&\underline{77.28}	&\textbf{72.46}&	64.37&	75.39&	\underline{74.35}&	11.88&	7.92&	56.68 \\
        &PL \cite{3dpl}& 53.01& 	42.17	& 39.41& 	60.59	& 60.24	& 72.29	& 71.60& 	73.32& 	75.04& 	69.02& 	64.03& 	73.49& 	70.05	& \textbf{12.22}& 	9.64& 	56.41 \\
        &DUA \cite{dua}& 51.81	&43.37&	41.13&	63.34&	62.99&	\textbf{73.67}&	\underline{73.15}&	73.32&	\textbf{77.97}&	\underline{71.94}&	\underline{66.78}&	\underline{75.56} &	\textbf{75.22}&	\underline{12.05}&	9.64	&58.13 \\
        &TENT \cite{tent}& \underline{53.87}	&44.23	&41.14&	63.86	&62.31	&\underline{72.98}&	72.46&	72.81	&77.11&	70.74	&\textbf{67.30}&	75.39	&73.67&	11.88&	\textbf{10.80}&	58.05 \\
        &SHOT \cite{shot}& 52.84	&44.41	&39.93	&65.06&	60.76&	72.12	&71.60&	72.98&	76.25	&69.02	&65.23	&72.46&	71.43&	11.88	&\underline{9.98}&	57.06 \\
        &BFTT3D \cite{bftt3d}& 33.51	&24.48&	39.93	&47.40&	44.79&	69.10&	69.79&	\textbf{74.31}	&74.83	&71.01&	54.86&	73.78&	71.70&	9.20&	7.81&	51.10 \\
        &CloudFixer \cite{cloudfixer}& \textbf{65.28}&	\textbf{63.72}&	\underline{46.53}&	\textbf{79.51}	&\textbf{78.30}&	65.45&	67.01&	69.27	&72.40&	65.62	&57.29&	68.92&	69.44&	9.38&	6.94&	\underline{59.01} \\
        \rowcolor{gray!12} &GSDTTA (ours)&  \underline{53.87}	&\underline{46.64}&	\textbf{69.88}&	\underline{74.35} &	\underline{63.51}&	72.81&	\underline{73.15}&	\underline{73.49}&	76.59	&71.26&	\underline{66.78}&	\textbf{75.73}&	72.98&	11.70&	9.81	&\textbf{60.84}\\
        \bottomrule  
    \end{tabular}}
        \caption{Classification accuracy (\%) across various distributional shifts in the ScanObjectNN-C dataset\cite{mate}. The results presented are based on three backbone models, each trained on the main split of the ScanObjectNN dataset \cite{scanobjectnn} and subsequently adapted to the OOD test set with a batch size of 32. Mean accuracy scores are reported with the highest values highlighted in bold and the second highest underlined.}
        \label{tab:scanobjectnnc}
        \vspace{-0.3cm}
\end{table*}

\subsection{Results}

\paragraph{ModelNet40-C.} Table~\ref{tab:modelnetc} provides a detailed performance comparison of various TTA methods on the ModelNet40-C~\cite{modelnetc} dataset, featuring 2D TTA methods like BN~\cite{bn}, PL~\cite{3dpl}, DUA~\cite{dua}, TENT~\cite{tent}, and SHOT~\cite{shot}, along with 3D-specific TTA methods such as BFTT3D~\cite{bftt3d} and CloudFixer~\cite{cloudfixer}. Our GSDTTA model achieves highest mean accuracy across all three backbones: 79.07\% (DGCNN), 82.63\% (CurveNet), and 82.51\% (PointNeXt) and maintains the highest or the second highest performance under most corruption types. Compared to SHOT~\cite{shot}, the best-performing 2DTTA method, GSDTTA achieves improvements of $1.71\%$, $1.39\%$, and $2.05\%$ on three backbones respectively, highlighting effectiveness of the special design of GSDTTA for 3D point cloud data. Additionally, GSDTTA achieves consistent improvements of $2.53\%$ (DGCNN), $4.72\%$ (CurveNet), and $6.47\%$ (PointNeXt) over the previous state-of-the-art 3DTTA method CloudFixer~\cite{cloudfixer}. Our method outperforms CloudFixer on 11 corruption types with DGCNN and CurveNet, and 12 types with PointNeXt. These improvements demonstrate the effectiveness of GSDTTA in adapting point clouds within the graph spectral domain at test time. The consistent gains across different backbones underscore GSDTTA’s adaptability and efficiency, establishing it as a robust solution for 3D test-time adaptation in point cloud classification under a variety of challenging distribution shifts.
\vspace{-0.3cm}

\paragraph{ScanObjectNN-C.} We conducted additional experiments on the challenging real-scanned point cloud dataset ScanObjectNN-C~\cite{mate} to further validate the effectiveness of GSDTTA. As shown in Table~\ref{tab:scanobjectnnc}, the source models for each backbone achieve relatively low classification accuracies, underscoring a significant distribution shift between ScanObjectNN-C and its clean counterpart, ScanObjectNN~\cite{scanobjectnn}. GSDTTA demonstrates notable improvements over existing methods across all tested backbones. Specifically, it surpasses CloudFixer, which operates in the spatial domain using diffusion models, with accuracy gains of $1.10\%$, $3.63\%$, and $1.83\%$ for three backbones respectively. It is worth nothing that CloudFixer outperforms our method on four basic noise-related corruptions (Uniform, Gaussian, Impulse, and Upsampling), with an average margin of $8.35\%$ across the three backbones. This is expected since CloudFixer specifically leverages diffusion models' denoising capabilities. For high-level semantic corruptions, GSDTTA demonstrates better average results on part dropping (Shear: +$4.68\%$, Cutout: +$3.60\%$) under three backbones. This shows that the overall mean improvement (+$2.19\%$) of GSDTTA comes from consistent performance across diverse corruptions. The better handling of semantic corruptions highlights the benefits of GSDTTA’s graph spectral approach, which effectively captures global structural features with a reduced number of parameters. This design enables GSDTTA to handle complex distribution shifts more efficiently than traditional spatial domain adaptations, making it particularly suitable for robust 3D test-time adaptation under real-world challenges.

\subsection{Ablation Study}
In this section, we take a closer look at the effects of components of GSDTTA, including the Graph Spectral Driven Point Shift (GSDPS) module for input adaptation, Graph Spectral Guided Model Adaptation (GSGMA) module for model adaptation, and eigenmap-guided self-training strategy. All experiments are conducted on the Scanobjectnn-C dataset with DGCNN~\cite{dgcnn} as backbone network. 
\vspace{-0.4cm}

\paragraph{Effectiveness of components in GSDTTA.}

We conduct an ablation study to evaluate the impact of each component in the GSDTTA framework. First, we remove the GSDPS module for input adaptation, denoting this variant as GSDTTA (w/o GSDPS), where only the original point cloud is processed by the model $f_\theta$ and adapted using the GSGMA module with the eigenmap-guided self-training strategy. As shown in Table~\ref{tab:ablation}, GSDTTA improves mean accuracy by $4.55\%$ over GSDTTA (w/o GSDPS) across 15 corruptions, demonstrating the effectiveness of GSDPS in adapting point clouds in the graph spectral domain. Next, we remove the GSGMA module, leaving only GSDPS adapt the input without model parameter updates, referred to as GSDTTA (w/o GSGMA). The full GSDTTA outperforms this variant by $3.48\%$, underscoring the importance of model adaptation with the eigenmap-guided self-training strategy. We also evaluate the impact of our outlier-aware graph by setting $\gamma=0$ in Eq.~\ref{eq:gamma}. Without this component, accuracy on background corruption drops dramatically from 69.54\% to 18.24\%, while average accuracy on the other 14 corruptions remains similar (61.44\% \textit{vs.} 61.28\%). This shows that the graphs are highly sensitive to outliers in background corruption, while the noises are easy to be remove.

\begin{figure}[t]
\caption{The classification accuracy of deep logits, spectral logits, and eigenmap-guided logits under some corruptions.}
\centering
\includegraphics[width=0.47\textwidth]{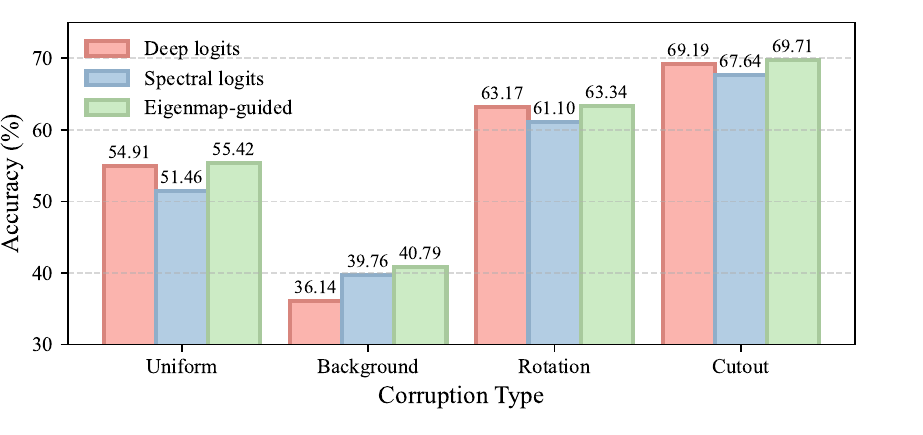}
\vspace{-0.4cm}
\label{fig:2b}
\end{figure}

To validate the eigenmap-guided self-training strategy, we generate three pseudo-label variants for some corruption types (Uniform, Background, Rotation, and Cutout). These pseudo-labels are obtained from clustering global deep descriptors, clustering global spectral descriptors, and the eigenmap-guided approach in Eqn.\ref{eq:combine}. As shown in Fig.~\ref{fig:2b}, eigenmap-guided pseudo-labels achieve superior performance on some corruptions with accuracy improvements compared to deep descriptor-based labels, validating our motivation presented in Sect.~\ref{subsec:motivation}. To further investigate the role of eigenmaps, we replace GSDTTA’s eigenmap-guided self-training strategy with a deep-feature-guided approach in Table \ref{tab:ablation}, which generates pseudo-labels solely from global deep descriptors. The $0.63\%$ performance degradation confirms the eigenmap's essential role as a domain-agnostic complement to source-specific features during initial adaptation.
\vspace{-0.3cm}

\begin{table}[!t]
\belowrulesep=0pt
\aboverulesep=0pt
\caption{Mean accuracy (\%) of variants of GSDTTA for point cloud classification on ScanObjectNN-C with DGCNN.}
\centering 
\begin{tabular}{cccc}
\toprule
GSDPS & GSGMA & EGSS & Mean \\
\toprule
 \ding{55}& \ding{55}& - & 55.72 \\
 \ding{55} & \checkmark &\checkmark & 57.28  \\
 \checkmark & \ding{55} & \checkmark& 58.35  \\
 \checkmark & \checkmark& \ding{55} & 61.20  \\
 \checkmark & \checkmark& \checkmark & \textbf{61.83}  \\
\bottomrule
\end{tabular}
\label{tab:ablation}
\vspace{-0.3cm}
\end{table}
\label{sec:experiment}

\begin{figure}[tbp]
    \caption{Sensitivity analysis of hyperparameters $\beta_1$, $\beta_2$, and $\beta_3$ on ScanObjectNN-C dataset under shear corruption with DGCNN.}
    \centering
    \begin{subfigure}[b]{0.155\textwidth}
        \includegraphics[width=\textwidth]{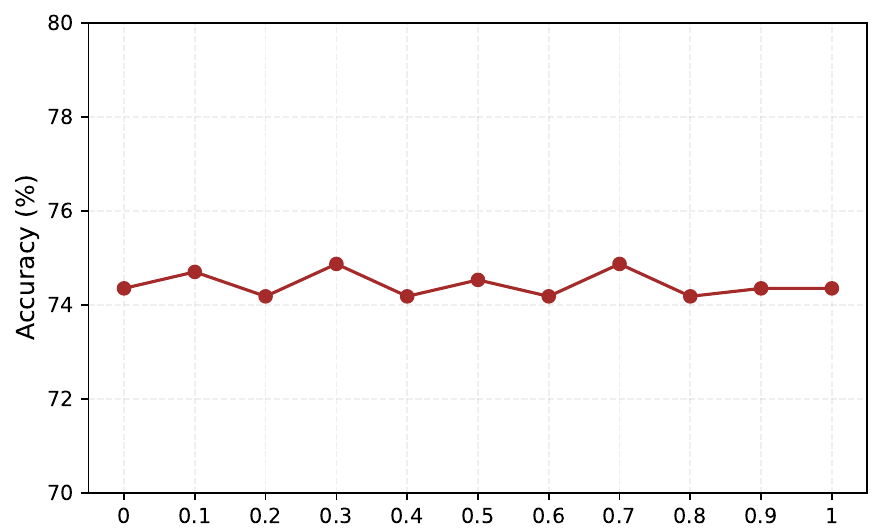}
        \caption{$\beta_1$}
        \label{fig:image2}
    \end{subfigure}
    \hfill
    \begin{subfigure}[b]{0.155\textwidth}
        \includegraphics[width=\textwidth]{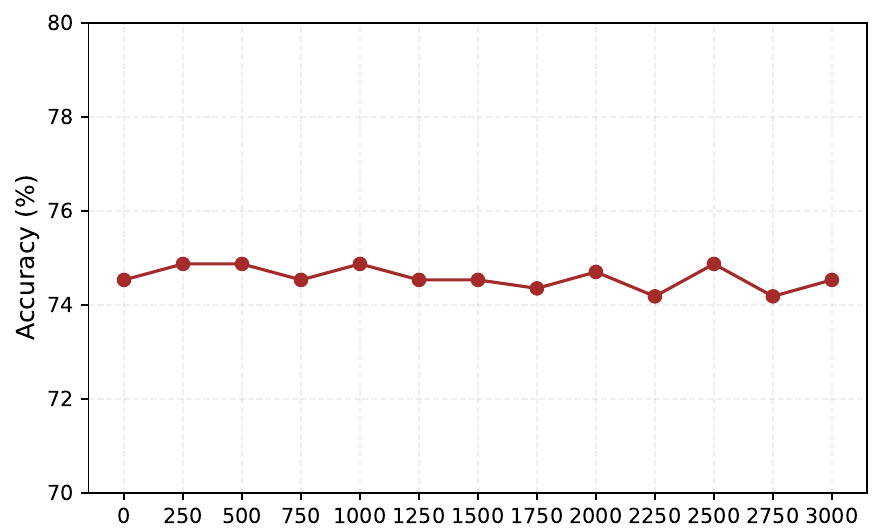}
        \caption{$\beta_2$}
        \label{fig:image3}
    \end{subfigure}
    \hfill
    \begin{subfigure}[b]{0.155\textwidth}
        \includegraphics[width=\textwidth]{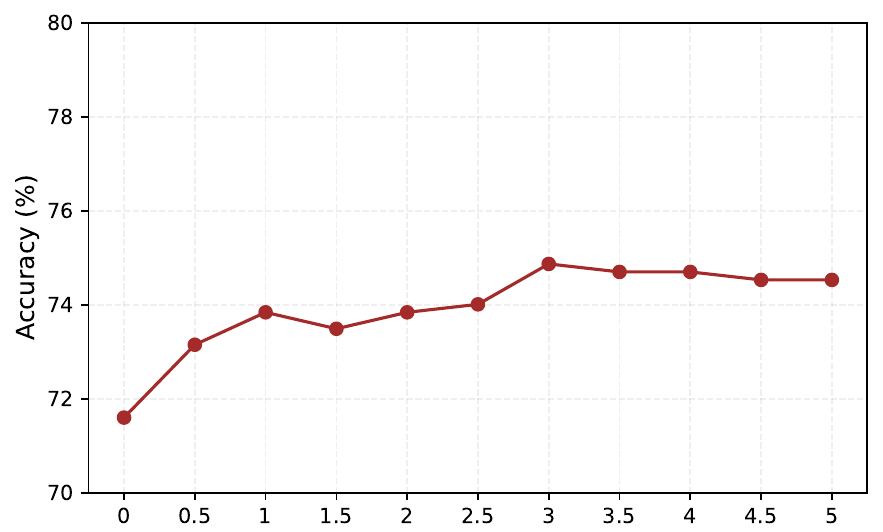}
        \caption{$\beta_3$}
        \label{fig:image4}
    \end{subfigure}
    \label{fig:sensitive}
    \vspace{-1cm}
\end{figure}

\paragraph{Sensitivity analysis on the hyperparameters.} We conduct an experiment on the ScanObjectNN-C dataset using DGCNN as the backbone model under shear corruption to analyze the sensitivity of weight factors $\beta_1$, $\beta_2$, and $\beta_3$ in the loss functions specified in Eqn.~\ref{eq:IA} and Eqn.~\ref{eq:MA}. For each experiment, other hyperparameters are fixed as described in Sect.~\ref{sec:implement}. As shown in Fig.~\ref{fig:sensitive}, the classification accuracy of GSDTTA remains stable with respect to $\beta_1$, $\beta_2$, and $\beta_3$ within the ranges of $[0, 1]$, $[0, 3000]$, and $[0, 5]$, respectively, with standard deviations of $0.27$, $0.25$, and $0.95$. This indicates that GSDTTA is not sensitive to these weight factors within the tested ranges.

\section{Related Work}
\label{sec:related_work}



\paragraph{Test-time adaptation.}
Test-time adaptation methods for 2D images have emerged as effective solutions to address challenges caused by domain shift, enabling models pre-trained on source domains to adapt dynamically to target domain data. For more details of 2D TTA methods, please refer to the supplementary materials. However, their applicability to 3D point cloud classification remains limited due to the unique challenges posed by irregular and unordered 3D data. MATE \cite{mate} addresses these challenges by employing masked autoencoders for self-supervised auxiliary tasks, enabling test-time adaptation to diverse point cloud distributions. BFTT3D \cite{bftt3d} takes a different approach by introducing a backpropagation-free adaptation model to mitigate error accumulation. Cloudfixer \cite{cloudfixer} leverages pre-trained diffusion models for point cloud denoising before adapting the classification model. Similarly, 3DD-TTA~\cite{3ddtta} employs a diffusion model to adapt target data to the source domain while maintaining frozen source model parameters. In contrast, GSDTTA departs from spatial-domain methods by leveraging the graph spectral domain for efficient and robust adaptation. By optimizing low-frequency spectral components where most of the point cloud's global structural information resides, GSDTTA significantly reduces the number of parameters requiring adaptation. 

\vspace{-0.3cm}
\paragraph{Graph spectral analysis for point clouds.}
Graph Spectral Analysis for Point Clouds. Just as frequency domain analysis enhances 2D vision models~\cite{phase-aware, 2dfrequency1, 2dfrequency2}, spectral methods for point clouds excel at analyzing intrinsic geometric structure. In point cloud matching, spectral analysis techniques extract features that capture the underlying structure of point clouds~\cite{spectral1, spectral2, spectral3, GPS, spectral5, spectral6, spectral7}. A key principle of the graph spectral domain is that low-frequency components preserve global structure, while high-frequency components capture finer details and noise. Leveraging this, spectral filters have been developed for denoising~\cite{HSP, GraphResampling}, while other methods manipulate Graph Fourier Transform (GFT) coefficients for attacks~\cite{gsda, SpectralAttack} or robust contrastive learning~\cite{GSPCon}. Building on these strengths, our method, GSDTTA, adapts both point clouds and model parameters in the graph spectral domain. We leverage the global structural information in low-frequency components to enable efficient and robust adaptation to distribution shifts, significantly improving performance in 3D classification tasks.
\section{Conclusion}
\label{sec:conclusion}
We proposed GSDTTA, a novel graph spectral domain test-time adaptation model that uses an eigenmap-guided self-training strategy. Extensive experiments validate its effectiveness on standard 3D-TTA benchmarks. While GSDTTA excels on these benchmarks, its scalability to large-scale point clouds is currently limited by the computational complexity of global spectral operations. Future work will address this by exploring unsupervised segmentation and multi-scale local spectral analysis to improve efficiency and reduce computational costs.

\vspace{0.2cm}
\noindent\textbf{Acknowledge} This work was supported in part by the China Postdoctoral Science Foundation under Grant Number 2025M771559, in part by the Postdoctoral Fellowship Program of CPSF under Grant Number GZB20250399, in part by the National Natural Science Foundation of China under Grants U22A2096 and 62036007, in part by Scientific and Technological Innovation Teams in Shaanxi Province under grant 2025RS-CXTD-011, in part by the Shaanxi Province Core Technology Research and Development Project under grant 2024QY2-GJHX-11, in part by the Fundamental Research Funds for the Central Universities under GrantQTZX23042, in part by the Young Talent Fund of Association for Science and Technology in Shaanxi China under Grant 20230121.
\\
{
    \small
    \bibliographystyle{unsrt}
    \bibliography{main}

\begin{thebibliography}{10}

\bibitem{pointnet}
Charles~R Qi, Hao Su, Kaichun Mo, and Leonidas~J Guibas.
\newblock Pointnet: Deep learning on point sets for 3d classification and
  segmentation.
\newblock In {\em CVPR}, pages 652--660, 2017.

\bibitem{dgcnn}
Yue Wang, Yongbin Sun, Ziwei Liu, Sanjay~E Sarma, Michael~M Bronstein, and
  Justin~M Solomon.
\newblock Dynamic graph cnn for learning on point clouds.
\newblock {\em ACM TOG}, 38(5):1--12, 2019.

\bibitem{curvenet}
Tiange Xiang, Chaoyi Zhang, Yang Song, Jianhui Yu, and Weidong Cai.
\newblock Walk in the cloud: Learning curves for point clouds shape analysis.
\newblock In {\em ICCV}, pages 915--924, 2021.

\bibitem{pointnet2}
Charles~Ruizhongtai Qi, Li~Yi, Hao Su, and Leonidas~J Guibas.
\newblock Pointnet++: Deep hierarchical feature learning on point sets in a
  metric space.
\newblock In {\em NeurIPS}, volume~30, 2017.

\bibitem{rscnn}
Yongcheng Liu, Bin Fan, Shiming Xiang, and Chunhong Pan.
\newblock Relation-shape convolutional neural network for point cloud analysis.
\newblock In {\em CVPR}, pages 8895--8904, 2019.

\bibitem{pct}
Meng-Hao Guo, Jun-Xiong Cai, Zheng-Ning Liu, Tai-Jiang Mu, Ralph~R Martin, and
  Shi-Min Hu.
\newblock Pct: Point cloud transformer.
\newblock {\em CVM}, pages 187--199, 2021.

\bibitem{simpleview}
Ankit Goyal, Hei Law, Bowei Liu, Alejandro Newell, and Jia Deng.
\newblock Revisiting point cloud shape classification with a simple and
  effective baseline.
\newblock In {\em ICML}, pages 3809--3820, 2021.

\bibitem{rpnet}
Haoxi Ran, Wei Zhuo, Jun Liu, and Li~Lu.
\newblock Learning inner-group relations on point clouds.
\newblock In {\em ICCV}, pages 15477--15487, 2021.

\bibitem{pointcnn}
Yangyan Li, Rui Bu, Mingchao Sun, Wei Wu, Xinhan Di, and Baoquan Chen.
\newblock Pointcnn: Convolution on x-transformed points.
\newblock In {\em NeurIPS}, volume~31, 2018.

\bibitem{pointmlp}
Xu~Ma, Can Qin, Haoxuan You, Haoxi Ran, and Yun Fu.
\newblock Rethinking network design and local geometry in point cloud: A simple
  residual mlp framework.
\newblock In {\em ICLR}, 2022.

\bibitem{spidercnn}
Yifan Xu, Tianqi Fan, Mingye Xu, Long Zeng, and Yu~Qiao.
\newblock Spidercnn: Deep learning on point sets with parameterized
  convolutional filters.
\newblock In {\em ECCV}, pages 87--102, 2018.

\bibitem{kpconv}
Hugues Thomas, Charles~R Qi, Jean-Emmanuel Deschaud, Beatriz Marcotegui,
  Fran{\c{c}}ois Goulette, and Leonidas~J Guibas.
\newblock Kpconv: Flexible and deformable convolution for point clouds.
\newblock In {\em ICCV}, pages 6411--6420, 2019.

\bibitem{pointnext}
Guocheng Qian, Yuchen Li, Houwen Peng, Jinjie Mai, Hasan Hammoud, Mohamed
  Elhoseiny, and Bernard Ghanem.
\newblock Pointnext: Revisiting pointnet++ with improved training and scaling
  strategies.
\newblock In {\em NeurIPS}, volume~35, pages 23192--23204, 2022.

\bibitem{modelnet40}
Zhirong Wu, Shuran Song, Aditya Khosla, Fisher Yu, Linguang Zhang, Xiaoou Tang,
  and Jianxiong Xiao.
\newblock 3d shapenets: A deep representation for volumetric shapes.
\newblock In {\em CVPR}, pages 1912--1920, 2015.

\bibitem{bn}
Steffen Schneider, Evgenia Rusak, Luisa Eck, Oliver Bringmann, Wieland Brendel,
  and Matthias Bethge.
\newblock Improving robustness against common corruptions by covariate shift
  adaptation.
\newblock In {\em NeurIPS}, volume~33, pages 11539--11551, 2020.

\bibitem{tent}
Dequan Wang, Evan Shelhamer, Shaoteng Liu, Bruno Olshausen, and Trevor Darrell.
\newblock Tent: Fully test-time adaptation by entropy minimization.
\newblock In {\em ICLR}, 2021.

\bibitem{t3a}
Yusuke Iwasawa and Yutaka Matsuo.
\newblock Test-time classifier adjustment module for model-agnostic domain
  generalization.
\newblock In {\em NeurIPS}, volume~34, pages 2427--2440, 2021.

\bibitem{tea}
Yige Yuan, Bingbing Xu, Liang Hou, Fei Sun, Huawei Shen, and Xueqi Cheng.
\newblock Tea: Test-time energy adaptation.
\newblock In {\em CVPR}, pages 23901--23911, 2024.

\bibitem{shot}
Jian Liang, Dapeng Hu, and Jiashi Feng.
\newblock Do we really need to access the source data? source hypothesis
  transfer for unsupervised domain adaptation.
\newblock In {\em ICML}, pages 6028--6039, 2020.

\bibitem{memo}
Marvin Zhang, Sergey Levine, and Chelsea Finn.
\newblock Memo: Test time robustness via adaptation and augmentation.
\newblock In {\em NeurIPS}, volume~35, pages 38629--38642, 2022.

\bibitem{IST}
Jing Ma.
\newblock Improved self-training for test-time adaptation.
\newblock In {\em CVPR}, pages 23701--23710, 2024.

\bibitem{dua}
M~Jehanzeb Mirza, Jakub Micorek, Horst Possegger, and Horst Bischof.
\newblock The norm must go on: Dynamic unsupervised domain adaptation by
  normalization.
\newblock In {\em CVPR}, pages 14765--14775, 2022.

\bibitem{dda}
Jin Gao, Jialing Zhang, Xihui Liu, Trevor Darrell, Evan Shelhamer, and Dequan
  Wang.
\newblock Back to the source: Diffusion-driven adaptation to test-time
  corruption.
\newblock In {\em CVPR}, pages 11786--11796, 2023.

\bibitem{FOA}
Shuaicheng Niu, Chunyan Miao, Guohao Chen, Pengcheng Wu, and Peilin Zhao.
\newblock Test-time model adaptation with only forward passes.
\newblock In {\em ICML}, 2024.

\bibitem{gda}
Yun-Yun Tsai, Fu-Chen Chen, Albert~YC Chen, Junfeng Yang, Che-Chun Su, Min Sun,
  and Cheng-Hao Kuo.
\newblock Gda: Generalized diffusion for robust test-time adaptation.
\newblock In {\em CVPR}, pages 23242--23251, 2024.

\bibitem{lame}
Malik Boudiaf, Romain Mueller, Ismail Ben~Ayed, and Luca Bertinetto.
\newblock Parameter-free online test-time adaptation.
\newblock In {\em CVPR}, pages 8344--8353, 2022.

\bibitem{tda}
Adilbek Karmanov, Dayan Guan, Shijian Lu, Abdulmotaleb El~Saddik, and Eric
  Xing.
\newblock Efficient test-time adaptation of vision-language models.
\newblock In {\em CVPR}, pages 14162--14171, 2024.

\bibitem{ttt}
Yu~Sun, Xiaolong Wang, Zhuang Liu, John Miller, Alexei Efros, and Moritz Hardt.
\newblock Test-time training with self-supervision for generalization under
  distribution shifts.
\newblock In {\em ICML}, pages 9229--9248, 2020.

\bibitem{ttt++}
Yuejiang Liu, Parth Kothari, Bastien~Germain van Delft, Baptiste Bellot-Gurlet,
  Taylor Mordan, and Alexandre Alahi.
\newblock Ttt++: When does self-supervised test-time training fail or thrive?
\newblock In {\em NeurIPS}, volume~34, pages 21808--21820, 2021.

\bibitem{mate}
M~Jehanzeb Mirza, Inkyu Shin, Wei Lin, Andreas Schriebl, Kunyang Sun, Jaesung
  Choe, Mateusz Kozinski, Horst Possegger, In~So Kweon, Kuk-Jin Yoon, et~al.
\newblock Mate: Masked autoencoders are online 3d test-time learners.
\newblock In {\em ICCV}, pages 16709--16718, 2023.

\bibitem{bftt3d}
Yanshuo Wang, Ali Cheraghian, Zeeshan Hayder, Jie Hong, Sameera Ramasinghe,
  Shafin Rahman, David Ahmedt-Aristizabal, Xuesong Li, Lars Petersson, and
  Mehrtash Harandi.
\newblock Backpropagation-free network for 3d test-time adaptation.
\newblock In {\em CVPR}, pages 23231--23241, 2024.

\bibitem{cloudfixer}
Hajin Shim, Changhun Kim, and Eunho Yang.
\newblock Cloudfixer: Test-time adaptation for 3d point clouds via
  diffusion-guided geometric transformation.
\newblock In {\em ECCV}, 2024.

\bibitem{3ddtta}
Hamidreza Dastmalchi, Aijun An, Ali Cheraghian, Shafin Rahman, and Sameera
  Ramasinghe.
\newblock Test-time adaptation of 3d point clouds via denoising diffusion
  models.
\newblock In {\em WACV}, 2025.

\bibitem{diffusion1}
Jonathan Ho, Ajay Jain, and Pieter Abbeel.
\newblock Denoising diffusion probabilistic models.
\newblock In {\em NeurIPS}, volume~33, pages 6840--6851, 2020.

\bibitem{diffusion2}
Jiaming Song, Chenlin Meng, and Stefano Ermon.
\newblock Denoising diffusion implicit models.
\newblock In {\em ICLR}, 2021.

\bibitem{diffusion3}
Yang Song, Jascha Sohl-Dickstein, Diederik~P Kingma, Abhishek Kumar, Stefano
  Ermon, and Ben Poole.
\newblock Score-based generative modeling through stochastic differential
  equations.
\newblock In {\em ICLR}, 2021.

\bibitem{spectral1}
Mathieu Aubry, Ulrich Schlickewei, and Daniel Cremers.
\newblock The wave kernel signature: A quantum mechanical approach to shape
  analysis.
\newblock In {\em ICCVW}, pages 1626--1633, 2011.

\bibitem{spectral2}
Jiaxi Hu and Jing Hua.
\newblock Salient spectral geometric features for shape matching and retrieval.
\newblock {\em VC}, 25:667--675, 2009.

\bibitem{spectral3}
Martin Reuter, Franz{-}Erich Wolter, and Niklas Peinecke.
\newblock Laplace–beltrami spectra as ‘shape-dna’ of surfaces and solids.
\newblock {\em CAD}, 38(4):342--366, 2006.

\bibitem{GPS}
Raif~M. Rustamov.
\newblock Laplace-beltrami eigenfunctions for deformation invariant shape
  representation.
\newblock In {\em SGP}, 2007.

\bibitem{spectral5}
Jian Sun, Maks Ovsjanikov, and Leonidas~J. Guibas.
\newblock A concise and provably informative multi-scale signature based on
  heat diffusion.
\newblock {\em CGF}, 28(5):1383--1392, 2009.

\bibitem{spectral6}
Yiqun Wang, Jianwei Guo, Dong-Ming Yan, Kai Wang, and Xiaopeng Zhang.
\newblock A robust local spectral descriptor for matching non-rigid shapes with
  incompatible shape structures.
\newblock In {\em CVPR}, pages 6231--6240, 2019.

\bibitem{spectral7}
Martin Weinmann, Boris Jutzi, and Cl{\'e}ment Mallet.
\newblock Semantic 3d scene interpretation: A framework combining optimal
  neighborhood size selection with relevant features.
\newblock {\em ISPRS Annals}, 2:181--188, 2014.

\bibitem{modelnetc}
Jiachen Sun, Qingzhao Zhang, Bhavya Kailkhura, Zhiding Yu, Chaowei Xiao,
  Z~Morley Mao, Ankit Goyal, Hei Law, Bowei Liu, Alejandro Newell, et~al.
\newblock Benchmarking robustness of 3d point cloud recognition against common
  corruptions.
\newblock In {\em ICML}, 2021.

\bibitem{3dpl}
Dong-Hyun Lee et~al.
\newblock Pseudo-label: The simple and efficient semi-supervised learning
  method for deep neural networks.
\newblock In {\em ICMLW}, volume~3, page 896, 2013.

\bibitem{adapt_p1}
Mohammad Fahes, Tuan-Hung Vu, Andrei Bursuc, Patrick P{\'e}rez, and Raoul
  de~Charette.
\newblock Poda: Prompt-driven zero-shot domain adaptation.
\newblock In {\em ICCV}, pages 18623--18633, 2023.

\bibitem{adapt_p2}
Jiachen Sun, Mark Ibrahim, Melissa Hall, Ivan Evtimov, Z.~Morley Mao, Cristian
  Canton~Ferrer, and Caner Hazirbas.
\newblock Vpa: Fully test-time visual prompt adaptation.
\newblock In {\em ACM MM}, pages 5796--5806, 2023.

\bibitem{im1}
Ryan Gomes, Andreas Krause, and Pietro Perona.
\newblock Discriminative clustering by regularized information maximization.
\newblock In {\em NeurIPS}, volume~23, 2010.

\bibitem{im2}
Yuan Shi and Fei Sha.
\newblock Information-theoretical learning of discriminative clusters for
  unsupervised domain adaptation.
\newblock In {\em ICML}, 2012.

\bibitem{scanobjectnn}
Mikaela~Angelina Uy, Quang-Hieu Pham, Binh-Son Hua, Thanh Nguyen, and Sai-Kit
  Yeung.
\newblock Revisiting point cloud classification: A new benchmark dataset and
  classification model on real-world data.
\newblock In {\em ICCV}, pages 1588--1597, 2019.

\bibitem{adamw}
Ilya Loshchilov and Frank Hutter.
\newblock Decoupled weight decay regularization.
\newblock In {\em ICLR}, 2019.

\bibitem{phase-aware}
Dawei Zhou, Nannan Wang, Heng Yang, Xinbo Gao, and Tongliang Liu.
\newblock Phase-aware adversarial defense for improving adversarial robustness.
\newblock In {\em International Conference on Machine Learning}, pages
  42724--42741. PMLR, 2023.

\bibitem{2dfrequency1}
Qidong Huang, Xiaoyi Dong, Dongdong Chen, Yinpeng Chen, Lu~Yuan, Gang Hua,
  Weiming Zhang, and Nenghai Yu.
\newblock Improving adversarial robustness of masked autoencoders via test-time
  frequency-domain prompting.
\newblock In {\em ICCV}, pages 1600--1610, 2023.

\bibitem{2dfrequency2}
Donghun Ryou, Inju Ha, Hyewon Yoo, Dongwan Kim, and Bohyung Han.
\newblock Robust image denoising through adversarial frequency mixup.
\newblock In {\em CVPR}, pages 2723--2732, 2024.

\bibitem{HSP}
Songyang Zhang, Shuguang Cui, and Zhi Ding.
\newblock Hypergraph spectral analysis and processing in 3d point cloud.
\newblock {\em IEEE TIP}, 30:1193--1206, 2020.

\bibitem{GraphResampling}
Siheng Chen, Dong Tian, Chen Feng, Anthony Vetro, and Jelena
  Kova{\v{c}}evi{\'c}.
\newblock Fast resampling of three-dimensional point clouds via graphs.
\newblock {\em IEEE TSP}, 66(3):666--681, 2017.

\bibitem{gsda}
Qianjiang Hu, Daizong Liu, and Wei Hu.
\newblock Exploring the devil in graph spectral domain for 3d point cloud
  attacks.
\newblock In {\em ECCV}, pages 229--248, 2022.

\bibitem{SpectralAttack}
Daizong Liu, Wei Hu, and Xin Li.
\newblock Point cloud attacks in graph spectral domain: When 3d geometry meets
  graph signal processing.
\newblock {\em IEEE TPAMI}, 2023.

\bibitem{GSPCon}
Yuehui Han, Jiaxin Chen, Jianjun Qian, and Jin Xie.
\newblock Graph spectral perturbation for 3d point cloud contrastive learning.
\newblock In {\em ACM MM}, pages 5389--5398, 2023.

\end{thebibliography}
}

\end{document}